\documentclass[bst/sn-basic]{sn-jnl}

\usepackage{graphicx}%
\usepackage{multirow}%
\usepackage{amsmath,amssymb,amsfonts}%
\usepackage[
        font={normal,small,sl,color=black!80},
        labelfont={normal,bf}
    ]{caption}
\usepackage{amsthm}%
\usepackage{mathrsfs}%
\usepackage[title]{appendix}%
\usepackage{xcolor}%
\usepackage{textcomp}%
\usepackage{manyfoot}%
\usepackage{booktabs}%
\usepackage{algorithm}%
\usepackage{algorithmicx}%
\usepackage{algpseudocode}%
\usepackage{listings}%
\usepackage{cite}
\usepackage{braket}
\usepackage{physics}
\definecolor{darkred}{rgb}{0.5,0.,0.}

\usepackage{orcidlink}

\usepackage{makecell}
\usepackage{natbib}



\begin{document}

\title[Article Title]{Quantum-inspired Embeddings Projection and Similarity Metrics for Representation Learning}

\author*[1,2]{\fnm{Ivan} \sur{Kankeu} \orcidlink{0009-0006-7470-4598}}

\author[2]{\fnm{Stefan Gerd} \sur{Fritsch}}

\author[3]{\fnm{Gunnar} \sur{Schönhoff}}

\author[3]{\fnm{Elie} \sur{Mounzer}}

\author[1,2]{\fnm{Paul} \sur{Lukowicz}}

\author[1,2,4]{\fnm{Maximilian} \sur{Kiefer-Emmanouilidis}}

\affil*[1]{\orgdiv{Department of Computer Science and Research Initiative QC-AI}, \orgname{RPTU Kaiserslautern-Landau}, \orgaddress{\city{Kaiserslautern}, \country{Germany}}}

\affil*[2]{\orgname{German Research Center for Artificial Intelligence (DFKI)}, \orgaddress{\city{Kaiserslautern}, \country{Germany}}}

\affil*[3]{\orgname{German Research Center for Artificial Intelligence (DFKI)}, \orgaddress{\city{Bremen}, \country{Germany}}}

\affil*[4]{\orgdiv{Department of Physics}, \orgname{RPTU Kaiserslautern-Landau}, \orgaddress{\city{Kaiserslautern}, \country{Germany}}}

\abstract{
Over the last decade, representation learning, which embeds complex information extracted from large amounts of data into dense vector spaces, has emerged as a key technique in machine learning. Among other applications, it has been a key building block for large language models and advanced computer vision systems based on contrastive learning. A core component of representation learning systems is the projection head, which maps the original embeddings into different, often compressed spaces, while preserving the similarity relationship between vectors.   

In this paper, we propose a quantum-inspired projection head that includes a corresponding quantum-inspired similarity metric. Specifically, we map classical embeddings onto quantum states in Hilbert space and introduce a quantum circuit-based projection head to reduce embedding dimensionality. To evaluate the effectiveness of this approach, we extended the BERT language model by integrating our projection head for embedding compression. We compared the performance of embeddings, which were compressed using our quantum-inspired projection head, with those compressed using a classical projection head on information retrieval tasks using the TREC 2019 and TREC 2020 Deep Learning benchmarks. The results demonstrate that our quantum-inspired method achieves competitive performance relative to the classical method while utilizing 32 times fewer parameters. Furthermore, when trained from scratch, it notably excels, particularly on smaller datasets. This work not only highlights the effectiveness of the quantum-inspired approach but also emphasizes the utility of efficient, ad hoc low-entanglement circuit simulations within neural networks as a powerful quantum-inspired technique.}


\maketitle

\section{Introduction}\label{sec1}
Representation learning involves learning a mapping from an input to a compact representation (i.e., a semantic space) that captures the inherent structure of the input data or concepts useful for downstream tasks \citep{bengio2013representation, lekhac2020contrastive}. The input is typically complex, high-dimensional data, such as videos, images, text, audio, or even multimodal, and is mapped to a feature vector, the so-called embedding, which is typically several orders of magnitude lower in dimension than the input. In many applications,  the generic representation is further processed by a projection head further optimize the features for a specific domain \citep{xue2024investigating}.  As an example, it has been determined that contrastive learning in computer vision is significantly improved by performing the similarity assessment not on the original embeddings but on the output of an additional trainable projection layer.  

Natural Language Processing (NLP), which is the specific application on which we demonstrate our concept of a quantum-inspired projection head, has been among the first areas to leverage representation learning. 
Early work focused on embedding single words based on the context in which they appear in a large corpus, such as in Word2Vec \citep{word2vec}, or through the co-occurrence of words, as in GloVe \citep{glove}. In contrast, recent models based on the Transformer architecture are capable of processing longer sequences. For example, BERT \citep{a73} can process sequences of up to 512 tokens\footnote{A token can be a word, subword, or character}, whereas OpenAI's state-of-the-art text embedding models can process inputs of up to 8192 tokens in length \citep{openai_embeddings}.
Sentence embeddings, serve as inputs for downstream tasks including text classification, question answering, and semantic textual similarity \citep{a73}.  Embeddings find extensive applications, including information retrieval systems for search engines \citep{a97} and, more recently, retrieval-augmented generation (RAG) for enhancing generative AI models with external knowledge \citep{a96}. In recent years, a tremendous amount of work has been done to improve the quality of the embeddings by using sophisticated and artistic training strategies that use increasingly intricate models \citep{a91,a92,a93,a94}. Models such as BERT, with millions of parameters, and large language models (LLMs) like LLaMA, which scale to billions of parameters, excel in producing high-quality embeddings \citep{a74,a95}.
However, the high-dimensional nature of these embeddings incurs substantial memory demands and computational overhead, namely for tasks involving information retrieval \citep{a84}.
To address the limitations posed by embedding dimensionality, various types of projection approaches have been proposed to compress the representation. Among these, dimension-preserving methods, such as embedding quantization, reduce memory usage by lowering the precision of vector components \citep{a90}. Dimensionality reduction methods like Principal Component Analysis (PCA), Kernel PCA, and more \citep{a81, a88, a89} have also been employed. These methods require complete access to the entire embedding set, which can be restrictive in real-world scenarios, where embeddings are generated on-the-fly. Neural network-based approaches tackle this by incorporating projection heads atop embedding models to compress embeddings into lower-dimensional representations during inference \citep{a84,a83}.

Quantum-inspired machine learning has recently gained traction alongside advancements in quantum computing, aiming to leverage principles of quantum mechanics and constructs from quantum computing to devise novel algorithms that operate on classical hardware, as noted by \cite{a98}. In this paper, we emphasize that quantum-inspired algorithms must be designed to ensure efficiency and scalability within classical computing environments, distinguishing them from complex quantum machine learning algorithms that can be to some extent simulated classically despite not being intrinsically efficient. The concept of drawing inspiration from quantum mechanics is not novel; it dates back to the 1990s when \cite{a99} explored concepts such as interference and superposition for computational tasks. However, with recent progress in both quantum computing and machine learning, quantum-inspired techniques have gained renewed interest, in model compression with tensor networks \citep{a206,a207} and particularly in adapting quantum computing formalisms for NLP applications, often with a focus on representation techniques \citep{a100, a101, a102, a103, a104}. For example, \cite{a102} proposed a method for constructing density matrices based on word embeddings to capture semantic information to solve question-answering tasks. Similarly, density matrices have been used to model semantic and sentiment nuances through projectors characterizing the semantic spaces of words \citep{a101, a103}. Additionally, we observe the emergence of quantum-inspired neural network architectures that incorporate quantum operations to simulate state evolution. \cite{a105} introduced a multimodal fusion approach that utilizes mixed-state representations of unimodal states, evolving these states through trainable unitary matrices, with observables used to interpret emotional states. Recently, \cite{a106} proposed a fine-tuning method for large language models inspired by quantum circuits to enable efficient high-rank adaptations via quantum operations. Leveraging the decomposition of arbitrary quantum circuits into sequences of single-qubit and two-qubit gates \citep{a107}, this approach provides a resource-efficient computational model, pushing the boundaries of classical machine learning with quantum-inspired innovation.

In this work, we introduce a theoretical framework, develop a quantum-inspired projection head architecture (including the associated quantum-inspired similarity metric) and validate it on the task of compressing the embedding space of the BERT language model. It leverages a quantum circuit-based projection head, consisting solely of single- and two-qubit gates, to compress BERT embeddings into lower-dimensional representations. Since this model is implemented on classical hardware, achieving both memory and computational efficiency is a primary focus. Our approach maps each classical BERT embedding to a quantum state within a Hilbert space, applying a sequence of quantum operations that culminate in projections to extract compressed information as a subsystem quantum state. This subsystem state is then used, in conjunction with a fidelity-based metric, to compute distances between embeddings. The BERT model is trained end-to-end with the parameterized quantum circuit in a Siamese network structure, which facilitates the learning of compressed, semantically meaningful sentence embeddings represented as quantum states in Hilbert space. Our contributions are as follows:

\begin{itemize}
	\item We introduce fidelity of fully separable states as an embedding distance metric.  
	\item We propose a quantum-inspired embedding projection approach with an efficient quantum circuit architecture, easily integrable as a projection head in embedding models.  
	\item We evaluate the concept on the specific problem of embedding compression for the BERT NLP system. We demonstrate superior performance in the passage ranking task of the quantum-inspired approach compared to its classical counterpart when the models are trained from scratch, particularly in data-scarce scenarios. 
\end{itemize}

The remainder of this paper is organized as follows: Section \ref{sec2} reviews the relevant literature and epitomizes prior works on representation learning in general, BERT training, and classical embedding compression techniques. Section \ref{sec3} details our proposed quantum-inspired approaches, including the theoretical foundation. Section \ref{sec4} presents the datasets, experimental setup, and materials used for evaluating our models. Sections \ref{sec5} \& \ref{sec6} analyze the experimental results and detail the ablation study. Finally, Section \ref{sec7} summarizes our findings and concludes.

\section{Related Works}\label{sec2}


\subsection{BERT}
\label{sec2.1}
BERT (Bidirectional Encoder Representations from Transformers) is a language representation model introduced by \citet{a73}. Building upon the transformer architecture \citep{a50}, BERT leverages large-scale unstructured textual data from the Internet to solve the language modeling task. Unlike earlier models that processed text in a unidirectional sequential manner \citep{a204,a205}, BERT uses a bidirectional approach, simultaneously considering both left-to-right and right-to-left contexts. This bidirectional approach, combined with the novel training strategy of masked language modeling, enabled BERT to achieve state-of-the-art results at the time of its release across a variety of NLP tasks such as language understanding, question answering, and common-sense inference \citep{a73}. To better understand the model, we will herein further elaborate on its architecture and its training procedures. 

\subsubsection{Architecture}
\label{sec2.1.1}
BERT is an encoder-only model based on the transformer architecture \citep{a50}, consisting solely of stacked encoder blocks and omitting the decoder component. Depending on the model variant (Base or Large), BERT can have up to 24 encoder blocks. Each block contains a multi-head attention layer \citep{a50} followed by a feed-forward layer, designed to compute dense embedding vectors that represent the contextual meaning of tokens (words or subwords) within a sequence. The embeddings from one block are passed as input to the next block, resulting in increasingly abstract and semantically enriched token representations. This ability to capture complex long-range dependencies between tokens is a key factor in BERT’s success and its widespread use in NLP tasks.

\subsubsection{Pre-training}\label{sec2.1.2}
BERT is pre-trained in an unsupervised manner using two main objectives. The first is Masked Language Modeling, where 15\% of the tokens in the input sequence are randomly selected for masking. Of these, 80\% are replaced with a special \texttt{[MASK]} token, while the remaining 20\% are either left unchanged or substituted with random words. The model’s goal is to predict the original tokens based on the surrounding context, thereby learning rich contextual embeddings. The second objective is Next Sentence Prediction, where sentence pairs are used as input, with some pairs being consecutive and others randomly paired. The model predicts whether the two sentences are consecutive, using the embedding of the special classification token (\texttt{[CLS]}) placed between them. This task encourages the model to understand the relationship between sentences.

\subsubsection{Fine-tuning}\label{sec2.1.3}

After the pre-training phase, BERT is typically fine-tuned in a supervised manner for specific tasks, allowing it to adapt the general knowledge it acquired -- such as understanding complex grammatical structures -- to effectively solve various downstream tasks. Common tasks include classification, question answering, and semantic textual similarity \citep{a73,a74}. In classification tasks, a classification layer is added on top of the \texttt{[CLS]} token's output to make predictions. For question answering, BERT processes a concatenation of the question and a text passage, separated by a \texttt{[SEP]} token, and is trained to predict the correct start and end positions of the answer within the passage. This is achieved by computing the scalar product of each token's representation with two trained vectors representing the start and end positions. In semantic textual similarity tasks, the goal is to measure how semantically similar two sentences are. A common approach is to use a Siamese architecture \citep{a74}, where two BERT instances, sharing the same weights, encode two sentences separately into embedding vectors. These vectors are then compared using a similarity measure, such as cosine similarity. A loss function, like contrastive loss, is used to bring embeddings of semantically similar sentences closer together and push dissimilar ones further apart, optimizing the model’s ability to capture semantic similarity.

\subsection{Embedding compression techniques}
\label{sec2.2}

Various techniques exist for reducing the embedding dimension and they can be classified into two categories: in-model and post-model techniques (Figure \ref{fig2.4}).

\vspace{0.75em}
\begin{center}
   \includegraphics[width=6cm, height=5cm]{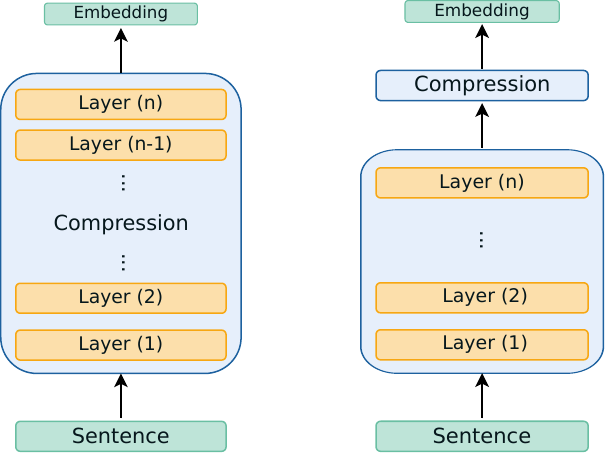}
   \captionof{figure}{Approaches to embedding compression.}
    \label{fig2.4}
\end{center}

\subsubsection{Post-model}
\label{sec2.2.1}

Post-model techniques, extensively studied in the literature, focus on dimensionality reduction methods applied after embeddings are generated by the model \citep{a76,a77,a78,a79,a80}. One such technique is Principal Component Analysis (PCA), a linear unsupervised method that projects high-dimensional embeddings onto a lower-dimensional subspace, retaining maximum variance by identifying key principal components \citep{a81}. Another technique is the Fast Fourier Transform (FFT), an efficient algorithm for converting data into the frequency domain. This method has been successfully applied to image compression \citep{a201,a202} and, more recently, to embedding compression by transforming embedding vectors into the frequency domain and truncating insignificant frequencies, thus preserving essential features while reducing storage and processing demands \citep{a82,d5,a79}. Autoencoders, a type of neural network for unsupervised learning, also serve as a dimensionality reduction tool. They achieve this by encoding input data into a compact lower-dimensional representation and reconstructing the original input, minimizing L2 reconstruction loss in the process \citep{a80}. Despite their effectiveness, post-model techniques face challenges such as poor generalization, the need to access the entire embedding set beforehand, and additional computational overhead.

\subsubsection{In-model}
\label{sec2.2.2}

In-model techniques address the previously mentioned issues by directly adapting the model architecture to produce embeddings of reduced dimensions. \cite{a83} propose a simple and efficient method to reduce sentence embeddings by projecting the \texttt{[CLS]} token embedding onto a lower-dimensional space using a fully connected projection layer followed by a non-linear activation function. Rather than training the BERT encoder and the projection layer in an end-to-end manner, they suggest a two-step training process. First, BERT is trained with a high-dimensional projection layer to obtain an optimal encoder. Then, the optimal encoder is fine-tuned alongside a pre-trained lower-dimensional projection layer. This approach aims to mitigate performance loss due to \textit{catastrophic forgetting} that can occur when training a pre-trained BERT model with a randomly initialized lower-dimensional projection layer.

\cite{a84} also propose reducing the dimensionality using a fully connected projection layer, but their method involves deriving the sentence embedding through mean pooling and training via distillation. In their approach, a larger BERT model, trained first, acts as the teacher. This model generates a high-dimensional embedding set, which is then compressed using PCA. A smaller BERT model, the student, is trained end-to-end with a projection layer. During training, the student’s embeddings are compared using mean squared error to the teacher’s embeddings, which have been reduced through PCA. The resulting loss is used to update the weights of both the projection layer and the encoder of the student model. The use of PCA enables homomorphic projective distillation.

\section{Proposed Methods}\label{sec3}
In this work, we introduce a novel quantum-inspired metric that leverages the fidelity of \textbf{fully separable} quantum states to measure similarity between embedding vectors encoded as quantum states. Additionally, we propose a quantum-inspired projection head that utilizes quantum state encodings to apply quantum operations, transforming embeddings into low-dimensional representations.

\subsection{Quantum-inspired metric learning}\label{sec3.1}
As briefly introduced in Section \ref{sec2.1.2}, BERT can be utilized for semantic textual similarity tasks. In these tasks, the model is trained to bring the embeddings of similar sentences closer together while pushing apart the embeddings of dissimilar sentences. The traditional metric resorted to for this training is cosine similarity, which measures the angle between two embedding vectors and determines whether they are pointing in the same direction. Here, we present a new quantum-inspired metric as a potential alternative to cosine similarity for metric learning using embeddings.

\subsubsection{Siamese BERT-networks}

Introduced by \cite{a74}, they provide an efficient method for training BERT to generate sentence embeddings. Before this approach, BERT was trained on sentence pairs to learn how to estimate sentence similarity, but it resulted in significant computational overhead \citep{a200}. With Siamese BERT, the model is trained to produce semantically meaningful representations of sentences, which can then be compared using a distance metric to determine semantic similarity.

Figure \ref{fig2.3} illustrates the standard BERT architecture during training for two sentences, where each undergoes the same processing steps. For each sentence, the sequence of token embeddings is computed using the same BERT model. A sentence-level embedding is then derived either through mean pooling (by averaging all token embeddings) or CLS pooling (by taking the embedding of the \texttt{[CLS]} token). The resulting embeddings are compared using cosine similarity.

In our approach, we merely replace the cosine similarity at the top of the architecture with a derived form of Uhlmann fidelity for pure states. By training BERT with this new metric, we push the model to adapt its embedding representations to align with the fidelity measure.

\vspace{0.75em}
\begin{center}
   \includegraphics[width=13cm, height=5.5cm]{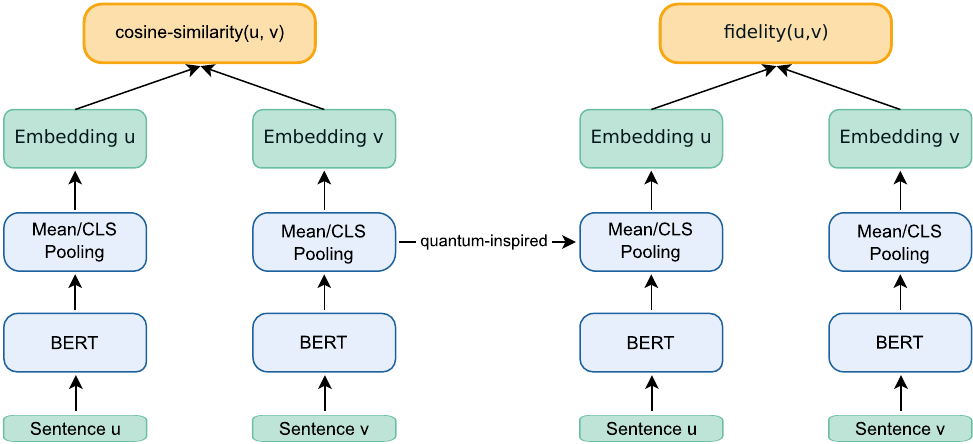}
   \captionof{figure}{BERT architecture at training.}
    \label{fig2.3}
\end{center}

\subsubsection{Quantum-inspired BERT (QiBERT)} 
QiBERT is built by mapping a pooled embedding of BERT to a quantum state in the Hilbert space. Given a pooled embedding $u=(u_1, u_2,\cdots, u_n)$, the corresponding quantum state in the Hilbert space \(\mathcal{H} = (\mathcal{C}^2)^{\otimes n}\) is constructed using Bloch sphere's polar coordinates as follows:
\begin{equation}\label{eq2.5}
  \begin{split}
    enc: \, &\mathcal{R}^n \to \mathcal{H} \\
         &u \mapsto \ket{u} = \ket{u_1} \otimes \ket{u_2} \otimes \cdots \otimes \ket{u_n}, \\
         &\text{with}\, \ket{u_i} =  \cos{\frac{\theta_i}{2}}\ket{0} + \sin{\frac{\theta_i}{2}}e^{i\phi_i}\ket{1},
  \end{split}
\end{equation}
where $\theta_i = tanh(u_i)\frac{\pi}{2} + \frac{\pi}{2}$ and $\phi_i = \pi$. The role of $tanh(\cdot)$ function is twofold, first, it normalizes the BERT logits within the range $[-1,1]$, ensuring $\theta_i$ falls within $[0, \pi]$, second, it fosters convergence during training, as noted by \cite{b3}, by pushing the average of each embedding over the training set closer to zero. To address the rapid saturation tendency of the hyperbolic tangent, the encoded vector can be either normalized or scaled down using a learnable regularization parameter, $\tau$, to modulate the degree of saturation. Alternatively, BERT's layers can also be fine-tuned to produce embeddings that align with the hyperbolic tangent function. Given another pooled embedding $v=(v_1, v_2,\cdots, v_n)$ with the corresponding quantum state $\ket{v} = enc(v)$, the semantic similarity is given by the fidelity as follows:
\begin{equation}
    \begin{split}
        fidelity(u,v) &= |\bra{u}\ket{v}|^2 = |(\bra{u_1} \otimes \bra{u_2} \otimes \cdots \otimes \bra{u_n})(\ket{v_1} \otimes \ket{v_2} \otimes \cdots \otimes \ket{v_n})|^2 \\
                      &= |\bra{u_1}\ket{v_1} \otimes \bra{u_2}\ket{v_2} \otimes \cdots \otimes \bra{u_n}\ket{v_n}|^2 \\
                      &= \prod_{i=1}^n|\bra{u_i}\ket{v_i}|^2.
    \end{split}
\end{equation}
The fidelity is classically tractable because the quantum state is \textbf{completely separable}. Instead of constructing the full quantum state vector with $2^n$ elements, $n$ fidelity scores are computed between pairs of 2-element vectors and then aggregated.

\subsection{Quantum-inspired embedding compression}\label{sec3.2}
Mapping classical embeddings to quantum states enables performing quantum operations on embedding quantum states before using the fidelity distance metric. Instead of using a fully connected layer, we propose employing quantum operations to create an efficient quantum compression head that can be trained end-to-end with BERT on a classical computer.

\subsubsection{Classical compression}
As discussed in Section \ref{sec2.2.2}, in-model classical compression is achieved by adding a projection layer on top of a $n$-layered BERT encoder, followed by a non-linear activation function. Figure \ref{fig2.5} illustrates this model architecture.

\begin{center}
\vspace{0.5cm}
   \includegraphics[angle=270,origin=c, width=7cm, height=4cm]{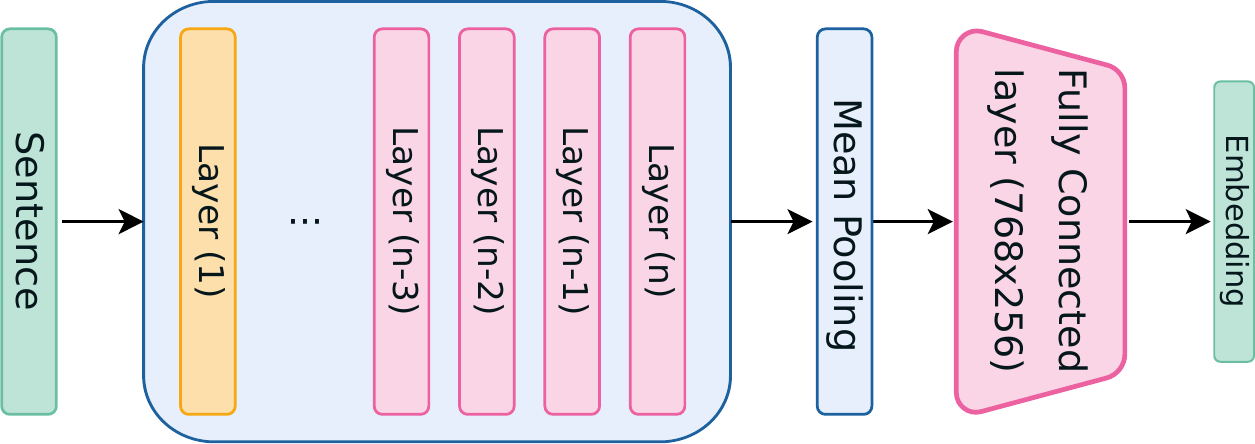}\vspace{-1.75cm}
   \captionof{figure}{Classical embedding compression for the base BERT model. Only the red components are tunable.}
   \label{fig2.5}
\end{center}
    
Given a n-layered BERT model \(\mathcal{B} = \tilde{\mathcal{L}}_{d\to d}^{(n-m,)}\circ\mathcal{L}_{\cdot\to d}^{(,n-m)}\), the compression algorithm can be mathematically formulated as follows:
\begin{equation}\label{eq2.4}
\mathcal{C}_c = \tilde{\mathcal{P}}_{d\to d'}\circ\mathcal{AP}\circ\tilde{\mathcal{L}}_{d\to d}^{(n-m,)}\circ\mathcal{L}_{\cdot\to d}^{(,n-m)}, \, \text{with}\, d' < d,
\end{equation}
where $m$ is the number of last layers trained, $\tilde{\mathcal{P}}$ denotes the projection layer with the activation function, $\mathcal{AP}$ is the average pooling, $\cdot\to\cdot$ indicates input and output dimensions as defined by \cite{a85}, $\cdot$ means any dimension, and $\circ$ is the composition operator. $\mathcal{L}^{(,j)}$ represents the composition of all encoder layers up to but not including the j-th layer, whilst $\mathcal{L}^{(j,)}$ is the composition of the j-th layer and all subsequent layers. This notation distinguishes between the trainable part of BERT ($\tilde{\mathcal{L}}_{d\to d}^{(n-m,)}$) and the frozen part ($\mathcal{L}_{\cdot\to d}^{(,n-m)}$). During training, the sentence embedding $e=(e_1,e_2,\cdots,e_j,\cdots,e_d)$ is pooled from the trainable part of BERT and passed to $\tilde{\mathcal{P}}$ for compression. For each dimension of the compressed output embedding $e'=(e'_1,e'_2,\cdots,e'_i,\cdots,e'_{d'})$, $\tilde{\mathcal{P}}$ calculates a linear combination of $e$:
\begin{equation}\label{eq2.1}
    e'_{i} = \sigma(\sum_{j=1}^d w_{ji}e_j + b_i),
\end{equation}
where $w_{ij}$ are the weights, $b_i$ the biases and $\sigma(\cdot)$ the activation function. Instead of considering all $d$ components of $e$ for every $e'_i$, only a unique subset of $\frac{d}{d'}$ components of $e$ can be used -- we assume, without loss of generality (w.l.o.g.), that $d$ is a multiple of $d'$. Therefore,
\begin{equation}\label{eq2.2}
\begin{split}
    &e'_{i} = \sigma(\sum_{j\in U_i} w_{ji}e_j + b_i), \\
    &\text{with}\, U_i \in \{U  \subseteq \{1,2,\cdots,d\} \, \big| \, |U|=\lfloor\frac{d}{d'}\rfloor \}.
\end{split}
\end{equation}
We conjecture that, in the optimal case, the compression produced by Equations \ref{eq2.1} and \ref{eq2.2} will be the same. Let $e'^{(1)},e'^{(2)}\in\mathcal{R}^{d'}$ be the compressed embeddings of $e^{(1)},e^{(2)}\in\mathcal{R}^{d}$, respectively, using Equation \ref{eq2.1}. Assuming w.l.o.g. that no activation function or bias is used and that all embeddings are normalized (i.e., $||\cdot||_2 = \sqrt{2(1-cos\_sim(\cdot))}$), we have
\begin{equation}
  e'^{(1)}_{i} =  \sum_{j=1}^{d} w_{ji}e_j^{(1)}, \, e'^{(2)}_{i} =  \sum_{j=1}^{d} w_{ji}e_j^{(2)}.
\end{equation}
Because of normalization, \(cosine\_similarity(e'^{(1)},e'^{(2)}) = 1 - \frac{||e'^{(1)}-e'^{(2)}||_2^2}{2}\). Thereby, the similarities between embeddings $\Delta' = ||e'^{(1)} - e'^{(2)}||_2^2$ and $\Delta = ||e^{(1)} - e^{(2)}||_2^2$, can be expressed as:
\begin{equation}
  \Delta' = \sum_{i=1}^{d'} \left(\sum_{j=1}^d w_{ji}(e_j^{(1)}-e_j^{(2)})\right)^2, \, \Delta = \sum_{j=1}^d (e_j^{(1)}-e_j^{(2)})^2.
\end{equation}
During training, the model should try to learn a projection matrix such that $\Delta'\approx\Delta$. In other words, each dimension of the compressed embedding should capture as much unique information as possible for faithful comparisons. Hence, $w_{ji}$ will act as attention scores to select the relevant information for the i-th dimension, such that:
\begin{equation}\label{eq2.3}
\begin{split}
    &\left(\sum_{j=1}^d w_{ji}(e_j^{(1)}-e_j^{(2)})\right)^2\approx \sum_{j\in U_i} (e_j^{(1)}-e_j^{(2)})^2,\\
    &\text{with}\, U_i \in \{U  \subseteq \{1,2,\cdots,d\} \, \big| \, |U|=\lfloor\frac{d}{d'}\rfloor\}.
\end{split}
\end{equation}
Therefore, in the optimal case, the model would need at most $\frac{d}{d'}$ parameters for every dimension, $w_{ji}=0$ for $j\not\in U_i$. The bottom line is that it should be possible to approximate the effect of a large projection matrix with fewer parameters. This rationale will be availed for our quantum-inspired compression approach.

\subsubsection{Quantum-inspired compression}\label{sec3.2.2}

Herein, we introduce the fundamental insights underlying quantum-inspired compression. The starting point is the preservation of ($\epsilon$-)distance between quantum states after compression, measured using the fidelity metric.

Let $\rho_n$ and $\sigma_n$ be quantum states of two training samples, and let $\{E_k\}_k$ be a set of positive operator-valued measurements (POVMs) such that \(\sum_k E_k = \mathbb{I}\). The fidelity bound between these states can be expressed as \(F(\rho_n, \sigma_n) \leq \left(\sum_k \sqrt{\mathrm{tr}(E_k\rho_n)\mathrm{tr}(E_k\sigma_n)}\right)^2\).
For ensembles of quantum states, this relationship generalizes to
\begin{equation}
    \sum_n F(\rho_n, \sigma_n) \leq \sum_n \left(\sum_k \sqrt{\mathrm{tr}(E_k\rho_n)\mathrm{tr}(E_k\sigma_n)}\right)^2.
\end{equation}
The compression process is defined by a unitary transformation $\textbf{\textit{U}}$ followed by partial measurement. For a 2-qubit system, this becomes \(\sum_n F(\rho_n, \sigma_n) \leq \sum_n \left(\sum_{k\in\{0,1\}} \sqrt{\mathrm{tr}(E_k\rho_n)\mathrm{tr}(E_k\sigma_n)}\right)^2\), with $E_k = \textbf{\textit{U}}^\dagger (\ket{k}\bra{k}\otimes \mathbb{I}) \textbf{\textit{U}}$.

The objective of this compression is to determine the optimal parameters $\theta$ for $\textbf{\textit{U}}$ that minimize distance errors, achieving
\begin{equation}\label{eq3.2.2.1}
    \sum_n F(\rho_n, \sigma_n) \approx^+ \min_{\theta} \sum_n \left(\sum_{k\in\{0,1\}} \sqrt{\mathrm{tr}(E_k(\theta)\rho_n)\mathrm{tr}(E_k(\theta)\sigma_n)}\right)^2.
\end{equation}
This optimization is equivalent to finding the optimal set of POVM operators $\{E_0(\theta),E_1(\theta)\}$ that maximally preserves the distance between quantum states after compression. 

Before moving to the circuit architecture of $\textbf{\textit{U}}$, we would like to note that the obtained classical probability distributions in Equation \ref{eq3.2.2.1} can in principle be re-encoded in a quantum state. Even though this might seem redundant at this point, we will make use of this in later sections where we conduct sequential concatenations. In this case, Equation \ref{eq3.2.2.1} extends to
\begin{equation}
    \min_{\theta} \sum_n \left(\sum_{k\in\{0,1\}} \sqrt{\mathrm{tr}(E_k(\theta)\rho_n)\mathrm{tr}(E_k(\theta)\sigma_n)}\right)^2 = \min_{\theta} \sum_n \left(\sum_{k\in\{0,1\}} \sqrt{p_{k,n}(\theta)q_{k,n}(\theta)}\right)^2
\end{equation}
with $p_{k,n}(\theta)$ and $q_{k,n}(\theta)$ corresponding to the probabilities.

\vspace{.75em}
\textbf{Circuit architecture of $\textbf{\textit{U}}$:} Instead of combining subsets of dimensions $U_i$ (Equation \ref{eq2.3}), we can combine the original dimensions in pairs of two for compression. To achieve this, the embedding needs to be mapped to fully separable quantum states, so that the compression is carried out by applying a two-qubit unitary operation to pairs of qubits. The whole process is depicted in Figure \ref{fig2.6}.

\vspace{0.75em}
\begin{center}
   \includegraphics[width=13cm, height=5cm]{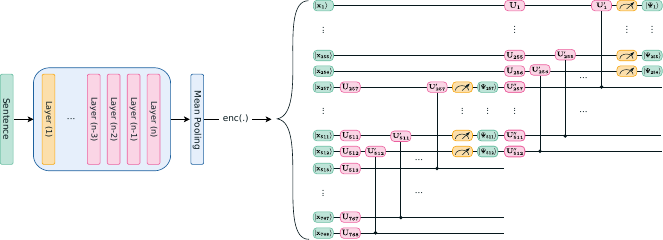}
   \captionof{figure}{Quantum-inspired embedding compression of base BERT. Only the red components are tunable; green components correspond to the different encoding methods ($\ket{\psi_i}$ writes back the measured probability distribution of the i-th qubit, and $enc(\cdot)$ is the data encoding as defined in Equation \ref{eq2.5}), and yellow components are fixed operations or frozen layers.}
    \label{fig2.6}
\end{center}
Given a n-layered BERT model \(\mathcal{B} = \tilde{\mathcal{L}}_{d\to d}^{(n-m,)}\circ\mathcal{L}_{\cdot\to d}^{(,n-m)}\) the compression process is formulated similarly to Equation \ref{eq2.4}, but this time the compression head $\tilde{\mathcal{Q}}$ is quantum-inspired:
\begin{equation}
\mathcal{C}_q = \tilde{\mathcal{Q}}_{d\to d'}\circ\mathcal{E}_{d\to d}\circ\mathcal{AP}\circ\tilde{\mathcal{L}}_{d\to d}^{(n-m,)}\circ\mathcal{L}_{\cdot\to d}^{(,n-m)}, \, \text{with}\, d' < d,
\end{equation}
where $\mathcal{E}$ maps an embedding $e$ to a quantum state $\ket{e}\in\mathcal{H}$ as defined in Equation \ref{eq2.5}. Just like in classical compression, the $m$ last layers and $\tilde{\mathcal{Q}}$ are trained, whilst the remaining model parameters are kept frozen. For a given high-dimension embedding $e=(e_1,e_2,\cdots,e_d)\in\mathcal{R}^d$ encoded as $\ket{e}=\bigotimes_{i=1}^d\ket{e_i}\in\mathcal{H}$, the quantum-inspired compression is defined in its general form as:
\begin{equation}
\begin{gathered}
        \tilde{\mathcal{Q}}_{d\to d'}\ket{e} = Q_{d'+1\to d'}^{(\cdot,\cdot)}\cdots Q_{d-1\to d-2}^{(\cdot,\cdot)}Q_{d\to d-1}^{(\cdot,\cdot)}\ket{e}, \\
        \text{with}\, Q_{d\to d-1}^{(i,j)} = \mathcal{M}_{i,j}CU_{i,j}(\mathbb{I}_1\otimes U(\theta_i)_i\otimes\mathbb{I}_{i+1}\otimes\cdots\otimes U(\theta_j)_j\otimes\cdots\otimes\mathbb{I}_d),
\end{gathered}
\end{equation}
where $Q_{d\to d-1}^{(i,j)}$ compresses the dimensions $i$ and $j$ into one by first applying two tunable universal single-qubit unitaries to the i-th qubit and j-th qubits. The universal unitary is constructed with a ZYZ unitary $U(\theta_i)_i=e^{i\alpha}R_Z(\beta)R_Y(\gamma)R_Z(\delta)$ and $\theta_i = (\alpha,\beta,\gamma,\delta)$ are learnable parameters. After applying the universal unitaries, a two-qubit controlled unitary $CU_{i,j}$ is applied with i-th qubit as the control and j-th qubit as the target -- note that this unitary is also universal and tunable. For simplicity, we assume w.l.o.g. that $j=i+1$ and the learned controlled unitary $CU_{i,i+1}=CNOT_{i,i+1}$. Thus, for a quantum state before applying the control unitary $\ket{e}^{(1)}=\ket{e_1}\otimes\cdots\ket{e_i}\otimes\ket{e_{i+1}}\otimes\cdots\ket{e_d}$ with $\ket{e_i}=a_i\ket{0} + b_i\ket{1}$ and $\ket{e_{i+1}}=a_{i+1}\ket{0} + b_{i+1}\ket{1}$, we have:
\begin{equation}
\begin{split}
  \ket{e}^{(2)}_{i,j} &= CNOT_{i,j}\ket{e}^{(1)}_{i,j} \\
           &= \ket{e_1}\otimes\cdots\otimes(
\mathrm{CNOT}(\ket{e_i}\otimes\ket{e_{i+1}}))\otimes\cdots\otimes\ket{e_d} \\
           &= \ket{e_1}\otimes\cdots\\
           &\,\,\,\,\,\, \otimes(a_ia_{i+1}\ket{00} + a_ib_{i+1}\ket{01} + b_ib_{i+1}\ket{10} + b_ia_{i+1}\ket{11})\otimes\\
           &\,\,\,\,\,\,\, \cdots\otimes\ket{e_d}.
\end{split}
\end{equation}
$\mathcal{M}_{i,j}$ involves repeated measurements of the $j$-th qubit in the computational basis -- herein, these measurements are obtained through simple projection, yielding the corresponding probabilities $p(0)$ and $p(1)$. This allows us to construct the probability distribution over the states $\ket{0}_j$, and $\ket{1}_j$. Indeed, a simple calculation shows that
\begin{equation}
\begin{split}
    p(0) &= |a_ia_{i+1}|^2 + |b_ib_{i+1}|^2 \\
    p(1) &= |a_ib_{i+1}|^2 + |b_ia_{i+1}|^2.
\end{split}
\end{equation}
Qubit $j$ is then reinitialized in the state
\begin{equation}
    \ket{e_j} = \sqrt{|a_ia_{i+1}|^2 + |b_ib_{i+1}|^2}\ket{0} + \sqrt{|a_ib_{i+1}|^2 + |b_ia_{i+1}|^2}\ket{1},
\end{equation}
whereas qubit $i$ is dropped. Thereby, the output quantum state has one less qubit and the $j$-th qubit state encodes a combination of information from the $i$-th and $j$-th dimensions. Thus, we have,
\begin{equation}
\begin{split}
    \ket{e'}_{i,j} &= \mathcal{M}_{i,j}\ket{e}^{(2)}_{i,j} \\
             &= \ket{e_1}\otimes\cdots\\
             &\,\,\,\,\,\, \otimes(\sqrt{|a_ia_{i+1}|^2 + |b_ib_{i+1}|^2}\ket{0} + \sqrt{|a_ib_{i+1}|^2 + |b_ia_{i+1}|^2}\ket{1})\otimes\\
             &\,\,\,\,\,\,\, \cdots\otimes\ket{e_{d-1}}.
\end{split}
\end{equation}
Clearly, $\ket{e'}$ is a compression of $\ket{e}$ from $d$ to $d-1$ using $Q_{d\to d-1}^{(i,j)}$. By applying $\tilde{\mathcal{Q}}_{d\to d'}$ repetitively, we can achieve an arbitrary $d'$-dimensional compression. 

For BERT, we carried out the compression as a cascaded two-step process (Figure \ref{fig2.6}). Initially, we compress the top $256$ dimensions with the middle $256$ dimensions. 

Following this through for all pairs of  the initial $512$ dimensions, as in Figure \ref{fig2.6} leads to the following state
\begin{equation}
    \ket{e'} = \bigotimes_{i = 1}^{256} \sqrt{|a_ia_{i+256}|^2 + |b_ib_{i+256}|^2}\ket{0} + \sqrt{|a_ib_{i+256}|^2 + |b_ia_{i+256}|^2}\ket{1}
\end{equation}

The resulting $512$-dimensional compressed embedding is then further reduced by compressing the top half with the bottom half, resulting in a densely compressed $256$-dimensional embedding. It is important to note that each unitary does not need to be applied sequentially to every qubit; instead, they can all be applied at once -- the computation is \textit{de facto} GPU-efficient. Nevertheless, the quantum-inspired compression may, in some cases, be slower than classical compression, as each \(\tilde{\mathcal{Q}}_{d \to d'}\) must be applied sequentially due to the measurement process at the end.

\section{Experiments}\label{sec4}
In this section, we describe the experiments conducted to evaluate and compare the performance differences between classical compression, quantum-inspired compression, and no compression. For further details on the experimental setup, see Appendix \ref{secA4}.

\subsection{Task \& Datasets}

To gauge the effectiveness of our quantum-inspired compression technique, we selected the asymmetric semantic search task. For training and evaluation, we employed three datasets derived from MS MARCO, a collection designed for deep learning in search tasks \citep{a75}. Specifically, we trained our models on an annotated subset of the MS MARCO Passage Ranking dataset and evaluated their performances on passage reranking task using the TREC $2019$ DL (TREC19) \citep{a86} and TREC 2020 DL (TREC20) \citep{a87} benchmarks (Table \ref{tab:A4.1}) with the popular NDCG@10 metric. NDCG evaluates the ranking model performance by considering both the relevance of results and their positions in the ordered list.

\subsection{Training settings}

We employed two base models for our experiments: ``google-bert/bert-base-uncased'' ($\mathrm{BERT_{base}}$) which is not fine-tuned and thus trained with a larger learning rate of 4e-5, and ``sentence-transformers/msmarco-distilbert-cos-v5'' ($\mathrm{BERT_{v5}}$) which is already fine-tuned on semantic textual similarity and therefore trained with a smaller learning rate of $2e-5$. Each base model was additionally trained using both a classical compression head (a fully connected layer followed by $\tanh(\cdot)$) and a quantum-inspired compression head (as defined in Section \ref{sec3.2.2}). For training, we established both session-specific settings and overarching global settings that apply consistently across all sessions.

Regarding global settings, the base models were fine-tuned by updating only the last four layers, while the compression models underwent training of both the compression heads and the last four layers. All models utilized a token input window of 256 and employed cross-entropy loss due to the weakly labeled data. The Adam optimizer was used for model optimization, and validation accuracy was assessed at the end of each epoch. The training set consists of 100K samples, while the validation set includes 10K samples. Early stopping was based on validation accuracy, with the model achieving the highest validation score selected for evaluation on benchmark datasets. The training was performed with a batch size of 64 on different GPUs (H100, A100-80GB, and RTXA6000).

We fine-tuned every model in a separate training session, with each being trained for $20$ runs. At every run, the seed value for ``torch(cuda)'', ``numpy'' and ``math'' libraries was increased by $1$ starting from $42$ and ending with $61$. The goal was to have the same initialization conditions, including, the same training batches and the same initial parameters.

\section{Result Analysis \& Discussion}\label{sec5}

We compare the performance of models with embedding compression to those without, as shown in Table \ref{tab5.1} (for detailed information on absolute model performance, see Tables \ref{tab:A8.1} and \ref{tab:A8.2}). 
Table \ref{tab5.1} is divided into two parts: the left part presents evaluation results on TREC19, while the right part presents results on TREC20. For each benchmark, the performance discrepancies between the base model ($\mathrm{BERT_{v5}}$ or $\mathrm{BERT_{base}}$) and its compression variants. The abbreviations BU, BT, BEC, and QBEC represent the base model untrained, base model trained, classical embedding compression model (256 dimensions), and quantum-inspired embedding compression model (256 dimensions), respectively.

\vspace{.75em}
\begin{tiny}
\begin{center}
\bgroup
\def\arraystretch{2}%
\setlength\tabcolsep{4.5pt}
\captionof{table}{Averages and standard errors of performance gaps (in percentage) across 20 runs between base models and their respective compression models. For instance, the first cell (QBEC,BU) indicates that the quantum-inspired compression model using $\mathrm{BERT_{v5}}$ underperforms the untrained base model $\mathrm{BERT_{v5}}$ by an average of -0.97\%, with an SE of 0.17\% on TREC19.}
\begin{tabular}{c | c c | c c | c c | c c }
 \toprule
 \multirow{3}{*}{\textbf{Models}} & \multicolumn{4}{c|}{ \makecell{TREC19\\\textbf{(Trained) Base Models}} } & \multicolumn{4}{c}{\makecell{TREC20\\\textbf{(Trained) Base Models}}} \\
 \cmidrule{2-9}
 & \multicolumn{2}{c|}{ $BERT_{v5}$ } & \multicolumn{2}{c|}{ $BERT_{base}$ } & \multicolumn{2}{c|}{ $BERT_{v5}$ } & \multicolumn{2}{c}{ $BERT_{base}$ } \\
 \cmidrule{2-9}
 & \textbf{BU} & \textbf{BT} & \textbf{BU} & \textbf{BT} & \textbf{BU} & \textbf{BT} & \textbf{BU} & \textbf{BT}\\
 \midrule
 
QBEC & -0.97$\pm$0.17 & -1.36$\pm$0.19  & \textcolor{gray}{90.89$\pm$0.8} & -0.04$\pm$0.41 & 0.24$\pm$0.27 & 0.15$\pm$0.23  
   & \textcolor{gray}{154.62$\pm$0.87} & 0.22$\pm$0.4 \\
\hline
BEC & -0.72$\pm$0.33 & -1.12$\pm$0.27 & \textcolor{gray}{89.13$\pm$0.72} & -0.95$\pm$0.42 & -0.6$\pm$0.31 & -0.68$\pm$0.31 & \textcolor{gray}{150.35$\pm$1.16} & -1.47$\pm$0.42\\
\hline
BT & 0.41$\pm$0.21 & - & \textcolor{gray}{90.98$\pm$0.56} & - & 0.09$\pm$0.14 & - & \textcolor{gray}{154.1$\pm$0.77} & -\\
\hline

\end{tabular}
\label{tab5.1}
\egroup
\end{center}
\end{tiny}
\vspace{.75em}

\textbf{Compression vs. base models:} As expected, Table \ref{tab5.1} reveals substantial performance improvements (gray values) between the untrained $\mathrm{BERT_{base}}$ (BU) and the trained models (QBEC, BEC, and BT). This is because $\mathrm{BERT_{base}}$ was never trained to produce sentence embeddings for semantic similarity, unlike $\mathrm{BERT_{v5}}$. After training $\mathrm{BERT_{base}}$ (BT), its quantum-inspired compression variant (QBEC) demonstrates comparable performance, with a minor drop on TREC19 (-0.04\%) and a slight gain on TREC20 (+0.22\%). For $\mathrm{BERT_{v5}}$, we see slight performance degradations (less than 1\%) on TREC19 for both the classical compression model (BEC, -0.72\%) and quantum-inspired compression model (QBEC, -0.97\%). However, on TREC20, QBEC marginally outperforms BU (+0.24\%). We also notice that the performance declines are somewhat more pronounced, and the gains slightly smaller, when compared to BT ($\mathrm{BERT_{v5}}$ trained on our dataset), indicating that training has a modest positive impact on $\mathrm{BERT_{v5}}$, unlike $\mathrm{BERT_{base}}$. The classical compression models (BEC) perform similarly to the quantum-inspired models (QBEC) when using $\mathrm{BERT_{v5}}$. However, with $\mathrm{BERT_{base}}$, the quantum-inspired models demonstrate superior compression performance.

The performance declines of QBEC and BEC on TREC19 are likely due to the smaller embedding size and fine-tuning effects, as noted in previous studies \citep{a83,a120}.  Conversely, the minor gains on TREC20, though unexpected, are consistent with findings in other research \citep{a76,a84}, where reduced embedding dimensions have occasionally slightly improved performance. We argue that the minor increases are attributable to the relatively small benchmark dataset size. In practice, smaller embeddings can increase the risk of false positives when dealing with large corpora, as observed by \cite{a120}.

\vspace{.75em}
\begin{tiny}
\begin{center}
\bgroup
\def\arraystretch{2}%
\setlength\tabcolsep{4.5pt}
\captionof{table}{Averages and standard errors of performance gaps (in percentage) across 20 runs between quantum-inspired and classical compression models for the backbone models, $BERT_{v5}$ and $BERT_{base}$.}
\begin{tabular}{c | c c | c c }
 \toprule
\multirow{2}{*}{\makecell{\textbf{Quantum-inspired} \\\textbf{Compression Models}}} & \multicolumn{2}{c|}{ \makecell{TREC19\\\textbf{Classical Compression Models}} } & \multicolumn{2}{c}{\makecell{TREC20\\\textbf{Classical Compression Models}}} \\
 \cmidrule{2-5}
 & \makecell{$BERT_{v5}$\\\textbf{BEC}} &  \makecell{$BERT_{base}$\\\textbf{BEC}} & \makecell{$BERT_{v5}$\\\textbf{BEC}} &  \makecell{$BERT_{base}$\\\textbf{BEC}}\\
 \midrule
QBEC & -0.23$\pm$0.28 & 0.94$\pm$0.39  & 0.85$\pm$0.34 & 1.73$\pm$0.45 \\
\hline
\end{tabular}
\label{tab5.2}
\egroup
\end{center}
\end{tiny}
\vspace{.75em}

\textbf{Quantum-inspired vs. classical compression models:} As previewed in Table \ref{tab5.1} and further highlighted in Table \ref{tab5.2}, the quantum-inspired approach (QBEC) demonstrates a slight performance decrease (-0.23\%) on TREC19, while achieving a more substantial performance increase (+0.85\%) on TREC20 when the fine-tuned model $BERT_{v5}$ is employed as the backbone. This suggests that the model has learned an equivalent, if not slightly superior, compression function with the quantum-inspired head compared to its classical counterpart. The performance disparity between the two approaches is even more pronounced when considering the $BERT_{base}$ model, which was trained for the first time on sentence representation, yielding nearly \textbf{+1\%} improvement on TREC19 and nearly \textbf{+2\%} improvement on TREC20. The competitive results are even more remarkable when accounting for the parameter counts associated with each compression method: $768 \times 256 + 256$ for the classical approach and $256 \times 6 \times 4$ for the quantum-inspired method, resulting in \textbf{32 times fewer} parameters required. Nonetheless, it remains uncertain whether these performances obtained can be ascribed solely to the compression head.

\vspace{.75em}

\begin{center}
   \includegraphics[width=13cm, height=4cm]{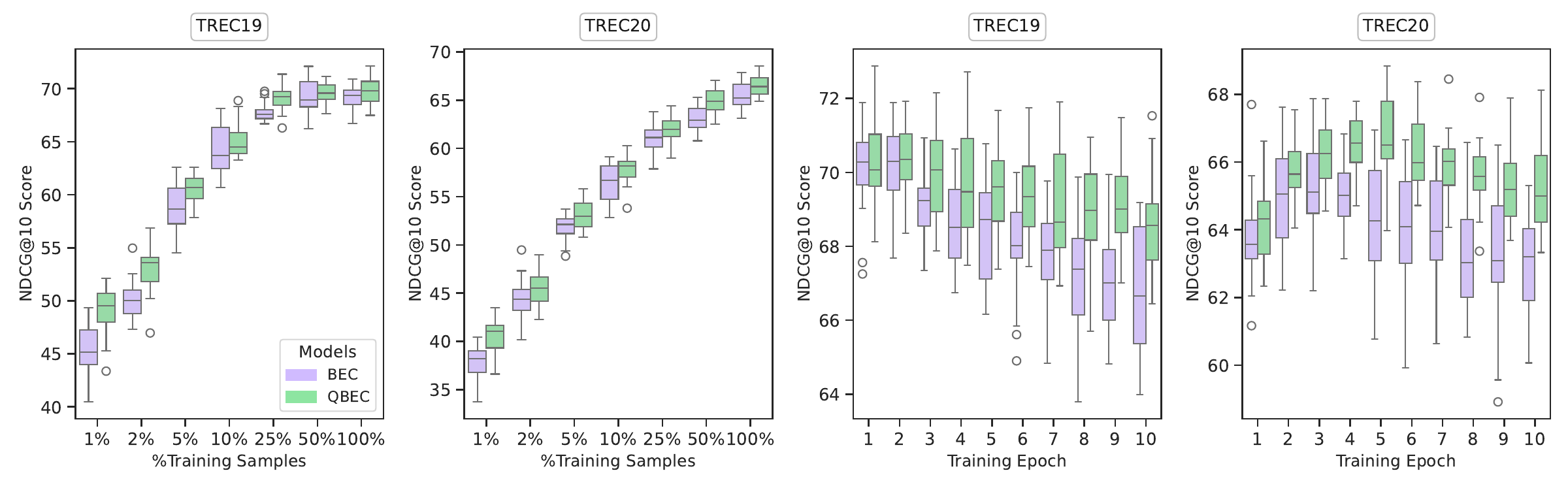}
   \captionof{figure}{Performance comparison of the classical and quantum-inspired compression models ($BERT_{base}$) on the TREC19 and TREC20 benchmarks. The two charts on the left illustrate the performance variation as a function of the training sample count, with 100\% corresponding to the use of 100K samples. The right two charts depict performance progression over training epochs using the full training sample set. The scores are scaled by 100.}
    \label{fig6.2}
\end{center}

\vspace{.75em}

\textbf{Training dataset size:} As shown in Table \ref{tab5.2}, the quantum-inspired model (QBEC) outperforms the classical model (BEC) when utilizing the $BERT_{base}$ backbone, which has not been pre-trained on sentence representations. To further investigate this performance advantage, we analyzed the effectiveness of the models across varying dataset sizes. The results indicate that QBEC consistently achieves superior performance on both benchmarks, even when trained from "scratch" on smaller datasets, as reflected in the medians presented in Figure \ref{fig6.2}. This trend is also corroborated by the mean values provided in Table \ref{tab:A8.3}. The performance disparity becomes increasingly pronounced as the training dataset size decreases, particularly on the TREC19 benchmark. For instance, while the performance gap between BEC and QBEC is approximately 1 point for the full dataset (100\%), this gap widens significantly to over 2.6 points for the smallest dataset (1\%) on both benchmarks (Table \ref{tab:A8.3}). These findings suggest that the quantum-inspired approach not only excels on large datasets but is particularly effective in data-scarce scenarios, which are frequently encountered in real-world applications.

\vspace{.75em}

\textbf{Convergence:} Further analysis of performance across training epochs reveals that QBEC outperforms BEC on median values at every epoch on the TREC20 benchmark, while exhibiting comparable performance on TREC19 (Figure \ref{fig6.2}). These observations are reinforced by performance averages, which indicate that QBEC achieves superior results as early as the first epoch across all benchmarks (Table \ref{tab:A8.4}). On TREC20, QBEC outpaces BEC by nearly 1 point at BEC’s peak performance (epoch 3). Notably, QBEC continues to improve beyond BEC’s peak, reaching its highest performance at epoch 5 with a 1.48-point advantage over BEC's peak value. Moreover, after reaching their respective peaks, BEC features a faster performance decline compared to QBEC on both benchmarks. This indicates that the quantum-inspired approach not only fosters convergence during training but also demonstrates a slight resistance to overfitting.

\section{Ablation Study}\label{sec6}

The ablation study examines which components contribute to the embedding compression model performance, the influence of the embedding dimension on performance, and the effects of utilizing a fidelity-based metric for learning.

\vspace{0.75em}
\begin{tiny}
\begin{center}
\bgroup
\def\arraystretch{1.5}%
\captionof{table}{Averages and standard errors of performance gaps (in percentage) across 20 runs between the base model $\mathrm{BERT_{v5}}$  and their respective frozen compression models.}
\begin{tabular}{c | c c | c c}
 \toprule
 \multirow{2}{*}{\textbf{Models}} & \multicolumn{2}{c|}{ \makecell{TREC19\\\textbf{(Trained) Base Model}} } & \multicolumn{2}{c}{\makecell{TREC20\\\textbf{(Trained) Base Model}}} \\
 \cmidrule{2-5}
 & \textbf{BU} & \textbf{BT} & \textbf{BU} & \textbf{BT} \\
 \midrule
 
QBFEC & -1.45$\pm$0.14 & -1.84$\pm$0.26 & -1.53$\pm$0.21 & -1.62$\pm$0.22\\
\hline
BFEC & -0.96$\pm$0.25 & -1.35$\pm$0.26 & -0.5$\pm$0.24 & -0.58$\pm$0.25\\
\hline

\end{tabular}
\label{tab6.1}
\egroup
\end{center}
\end{tiny}
\vspace{.75em}

\textbf{Quantum-inspired vs. classical head:} To assess the impact of the additional compression head on top of BERT, we retrained the models with all BERT parameters frozen. This produced two models: QBFEC, where only the unitary parameters of QBEC were trained, and BFEC, where only the fully connected projection layer of BEC was trained. 
A comparison between Tables \ref{tab5.1} and \ref{tab6.1} reveals an overall performance decline for both classical and quantum-inspired approaches when the BERT encoder layers are frozen, with the exception of the classical approach on TREC20. This finding suggests that fine-tuning BERT's last layers contributed to the improved performance of both QBEC and BEC. Furthermore, the performance decline is more pronounced for the quantum-inspired approach, which now underperforms the classical approach by -0.48$\pm$0.31\% on TREC19 and -1.03$\pm$0.31\% on TREC20. These results lead to two key conclusions: (1) the quantum-inspired head may have a lower representational power than the classical head, and (2) the performance of QBEC and BEC is not solely due to their compression heads but also to the joint training with BERT's final layers. Overall, these observations suggest that fine-tuning BERT's layers bolsters the capacity of compression models to approximate complex functions, beyond what the compression heads alone could achieve.

\vspace{.75em}
\begin{center}
   \includegraphics[width=13cm, height=4.8cm]{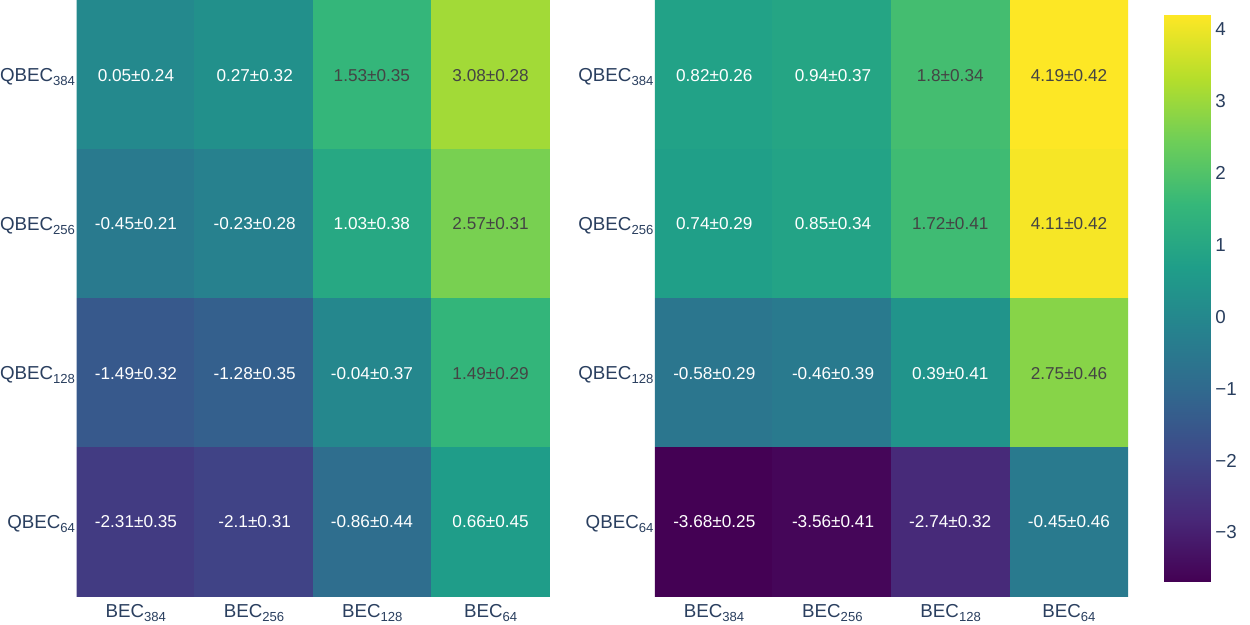}
   \captionof{figure}{Averages and standard errors of performance gaps (in percentage) across 20 runs between quantum-inspired and classical compression models (64, 128, 256, and 384 dimensions) for the backbone model $BERT_{v5}$. The heatmaps on the left and right sides display results for the TREC19 and TREC20 benchmarks, respectively.}
    \label{fig6.1}
\end{center}

\textbf{Effect of the embedding size:} To analyze the impact of embedding size, we compared the performance of quantum-inspired and classical compression models across various output dimensions: 64, 128, 256, and 384. The quantum-inspired compression model with dimension $d$ ($\mathrm{QBEC}_{d}$) retains the architecture outlined in Section \ref{sec3.2.2}, where information is sequentially processed from the final $d$ dimensions down to the initial $d$ dimensions. The final embedding is taken from this reduced set, specifically the range $[0, d]$. 

Figure \ref{fig6.1} heatmaps display performance discrepancies between quantum-inspired and classical compression models for each number of output dimensions. While no substantial performance differences are observed between quantum-inspired and classical models for the same dimension count, averaging the two diagonals over both benchmarks yields a trend of slight but consistent performance declines (0.44\%, 0.31\%, 0.18\%, and 0.11\% for sizes 384, 256, 128, and 64, respectively) in quantum-inspired compression as embedding size decreases. This decline is likely due to the compression head depth, which increases with lower output dimensions (e.g., one layer at 384 dimensions, two at 256, and eleven at 64). Higher depth may hinder effective information flow to the first 64 dimensions, whereas, with 384 dimensions, information can be transferred directly from the last to the first half of the embedding. Moreover, due to the approximate nature of the equation.
\ref{eq3.2.2.1}, small deviations are to be expected with every compression step. 

An additional trend is that model performance consistently improves with larger embedding sizes, as seen across each row and column of the heatmaps. This pattern suggests that both the architecture of the compression head and the embedding size impact compression quality.

\vspace{.75em}

\begin{tiny}
\begin{center}
\bgroup
\def\arraystretch{1.5}%
\captionof{table}{Averages and standard errors of performance gaps (in percentage) across 20 runs between fidelity-based base models and other models.}
\begin{tabular}{c | c c | c c}
 \toprule
 \multirow{2}{*}{\textbf{Models}} & \multicolumn{2}{c|}{ \makecell{TREC19\\\textbf{Fidelity-based Base Model}} } & \multicolumn{2}{c}{\makecell{TREC20\\\textbf{Fidelity-based Base Model}}} \\
 \cmidrule{2-5}
 & \makecell{$BERT_{v5}$\\\textbf{QBT}} &  \makecell{$BERT_{base}$\\\textbf{QBT}} & \makecell{$BERT_{v5}$\\\textbf{QBT}} &  \makecell{$BERT_{base}$\\\textbf{QBT}}\\
 \midrule

QBEC & -1.4$\pm$0.19 & -1.31$\pm$0.29 & -0.78$\pm$0.25 & -0.94$\pm$0.39\\
\hline
BEC & -1.16$\pm$0.33 & -2.21$\pm$0.31 & -1.61$\pm$0.3 & -2.6$\pm$0.46\\
\hline
BT & \textbf{-0.03$\pm$0.22} & \textbf{-1.24$\pm$0.4} & \textbf{-0.93$\pm$0.14} & \textbf{-1.14$\pm$0.32}\\
\hline

\end{tabular}
\label{tab6.2}
\egroup
\end{center}
\end{tiny}

\vspace{.75em}

\textbf{Fidelity-based metric:} To evaluate the effect of employing the fidelity of fully separable states as an embedding distance metric in place of cosine similarity, we retrained the base models ($\mathrm{BERT_{base}}$ and $\mathrm{BERT_{v5}}$) using the fidelity-based metric and compared their performances with those of other models. Table \ref{tab6.2} epitomizes the evaluation results, where QBT and BT represent the base models trained with fidelity and cosine similarity, respectively. The results indicate that the $\mathrm{BERT_{v5}}$ model trained with cosine similarity underperformed by \textbf{-0.03\%} on TREC19 and \textbf{-0.93\%} on TREC20 compared to its fidelity-based counterpart. This performance gap is even more accentuated for $\mathrm{BERT_{base}}$, which was not initially trained for sentence representation, with the cosine similarity-based model showing a \textbf{-1.24\%} and \textbf{-1.14\%} deficit on TREC19 and TREC20, respectively. These findings suggest that the use of fidelity alone enhances the quality of sentence representations, thereby improving benchmark performance. Therefore, it is likely that the fidelity metric could also positively influence the training of quantum-inspired embedding compression models. There is a notable proximity (see Appendix \ref{secA9}) between the fidelity and Manhattan distance metrics, which \cite{a119} identified as suitable for high-dimensional data to alleviate the \textit{curse of dimensionality}.

\section{Conclusion}\label{sec7}

In this paper, we have proposed a quantum-inspired projection head for representation learning in systems and evaluated it on the specific task of compressing the embedding space of the BERT language model. We show that the classical method of embedding compression using a fully connected layer can be effectively approximated by selectively using subsets of input embedding dimensions for each output dimension. Building on this, we introduced a novel quantum-inspired technique that leverages controlled unitary operations to combine embedding dimensions in pairs, enabling efficient information transfer across the embedding for compression purposes. Empirical evaluations showed that our approach delivers competitive performance on passage-ranking tasks with 32 times fewer parameters than its classical pendant. Remarkably, when trained from scratch, it not only meets but also exceeds performance, particularly on smaller datasets.

To enable this quantum-inspired projection, we also proposed a quantum encoding strategy that maps classical embeddings to quantum states, allowing the use of quantum operations and a constrained fidelity metric as a distance measure between embeddings. This metric was found to enhance the performance of BERT-based models when used for metric learning. We hope our work, due to its simplicity, paves the way for broader adoption of quantum-inspired techniques in machine learning. While it has been evaluated on an NLP problem the method the design is generic and should be easily adaptable to a broad range of other representation learning problems. 

\paragraph{Acknowledgements} We gratefully acknowledge financial support from the Quantum Initiative Rhineland-Palatinate QUIP and the Research Initiative Quantum Computing for AI (QC-AI). We acknowledge support from the German Federal Ministry of Research and Education (BMBF) through the project Q3-UP! under project number 13 N 15 779 administered by the technology center of the Association of German Engineers (VDI) and by the German Federal Ministry of Economic Affairs and Climate Action (BMWK) through the project QuDA-KI under the project numbers 50RA2206A administered by the German Aerospace Center (DLR).

\paragraph{Author Contributions} I.K., S.G.F., E.M., P.L., and M.K-E. conducted the research. I.K. drafted the initial manuscript and, together with E.M., created the figures. I.K., S.G.F., and M.K-E. optimized the code on GPU, while I.K. and S.G.F. generated the results. P.L. and M.K-E. supervised the project. All authors contributed to the manuscript’s text and have read, reviewed and approved the final manuscript.

\paragraph{Data and Code availability}
The source code used in this manuscript is available at \href{https://github.com/ivpb/qiepsm}{https://github.com/ivpb/qiepsm}.

\section*{Declarations}
\paragraph{Conflict of Interest} The authors declare no competing interests.

\bibliographystyle{bst/sn-basic}

\begin{appendices}

\section{Experiment Details}\label{secA4}

\subsection{Task \& Datasets}

The asymmetric semantic search involves using a short query (a few keywords) to find a longer text passage that answers the query. In semantic search, a corpus is typically embedded using an embedding model -- in this case, BERT -- and stored in a vector database. Upon receiving a query, the retriever searches the database, computes the similarity between each stored embedding and the query embedding, and returns the top $k$ most similar items. 

\defcitealias{d3}{Sentence Transformers}
The training dataset was generated using labels from \citetalias{d3}. The labels were created by scoring 160 million (query, passage) pairs using a cross-encoder. The training samples are structured as $(q, pos, neg_1,\cdots, neg_5)$, where $q$ is the query, $pos$ is a positive passage relevant to the query, and $neg_i$ are hard negatives. Hard negatives are selected based on the condition that their score difference with the positive passages is greater than 3, ensuring they are truly less relevant than the positive passages. The objective of the model during training is to predict the most relevant passage.

TREC 2019 DL (TREC19) \citep{a86} and TREC 2020 DL (TREC20) \citep{a87} are both benchmarks employed for evaluation. They were derived from the passage retrieval dataset by selecting 43 and 54 queries, respectively and manually labeling the associated passages on a scale from 0 to 3, with 0 being irrelevant and 3 being perfectly relevant. During evaluation, the model goal is to rerank the passage set for each query so that the results align with the relevance order. The standard metric associated with these benchmarks is the NDCG metric. The statistics for the different datasets used can be found in Table \ref{tab:A4.1}. 

\begin{small}
\begin{center}
\captionof{table}{Statistics of different datasets used}
\begin{tabular}{c c c}
 \toprule
 \textbf{Datasets} & \textbf{\#Query} & \textbf{\#Passage} \\
 \midrule
  Training set & 494,560 & 2,153,951 \\ 
 \hline
 TREC19 & 43 &  9,260 \\ 
 \hline
 TREC20 & 54 &  11,386 \\ 
 \hline
\end{tabular}
\label{tab:A4.1}
\end{center}
\end{small}
\subsection{Evaluation metric} 
To assess the quality of the trained model, we utilized the NDCG metric commonly employed in information retrieval. NDCG (Normalized Discounted Cumulative Gain) measures ranking quality in information retrieval. It is the ratio of DCG (Discounted Cumulative Gain) to IDCG (Ideal Discounted Cumulative Gain) which represents a perfect ranking. NDCG is typically calculated for the top $k$ items in the result list: \(NDCG@K = \frac{DCG@K}{IDCG@K}\), where \(DCG@K = \sum_{i=1}^k\frac{rel_i}{\log_2(i+1)}\). Here, $rel_i$ is the relevance score of the i-th item, and \(\log_2(i+1)\) serves as a penalty for incorrect rankings. Thus, a highly relevant item that is ranked incorrectly will have a greater impact on the final score than a less relevant item. For our experiments, we set $k=10$, as recommended for the TREC benchmark datasets. We used the Python library ``torchmetrics'' to calculate NDCG scores in our implementation.

\section{Fidelity \& Manhattan Metric}\label{secA9}

To highlight the proximity between the fidelity and Manhattan distance metrics, consider two embedding vectors encoded using Equation \ref{eq2.5}, $\ket{u} = \bigotimes_{i=1}^n\ket{u_i}$ and $\ket{v} = \bigotimes_{i=1}^n\ket{v_i}$, where $\ket{u_i} = \cos{\frac{\theta_i^u}{2}}\ket{0} + \sin{\frac{\theta_i^u}{2}}\ket{1}$ and $\ket{v_i} = \cos{\frac{\theta_i^v}{2}}\ket{0} + \sin{\frac{\theta_i^v}{2}}\ket{1}$. The fidelity between these vectors is given by
\begin{equation}
    \begin{split}
        fidelity(u,v) &= \prod_{i=1}^n |\bra{u_i}\ket{v_i}|^2 \\
                      &= \prod_{i=1}^n |\cos{\theta_i^u}\cos{\theta_i^v} + \sin{\theta_i^u}\sin{\theta_i^v}|^2 \\
                      &= \prod_{i=1}^n |\cos{(\theta_i^u - \theta_i^v)}|^2. \\
    \end{split}
\end{equation}
For non-zero fidelity values, the logarithm is strictly monotonic and increasing, making it an order-preserving transformation for fidelity:
\begin{equation}
fidelity(u,v) \sim \frac{1}{2}\log(fidelity(u,v)) = \sum_{i=1}^n \log(\cos{(|\theta_i^u - \theta_i^v|)}).
\end{equation}
The latter formula bears a close resemblance to the L1 (Manhattan) distance metric (\(\sum_{i=1}^n (|\theta_i^u - \theta_i^v|)\)).

\section{Evaluation Results}\label{secA8}

\begin{tiny}
\begin{center}
\bgroup
\def\arraystretch{1.5}%
\captionof{table}{Averages and standard errors of performance scores (in NDCG@10) across 20 runs for models using $\mathrm{BERT}_{v5}$ as backbone and base models. The scores are scaled by 100.}
\begin{tabular}{c | c c}
 \toprule
 \textbf{Models} & \textbf{TREC19} & \textbf{TREC20} \\
 
 \midrule
& \multicolumn{2}{c}{\textbf{Quantum-inspired Embedding Compression Models}} \\
 \midrule
 
$\mathrm{QBEC}_{384}$ & 73.01$\pm$0.1 & 67.37$\pm$0.11\\
\hline
QBEC or $\mathrm{QBEC}_{256}$ & 72.65$\pm$0.12 & 67.32$\pm$0.18\\
\hline
$\mathrm{QBEC}_{128}$ & 71.88$\pm$0.14 & 66.44$\pm$0.2\\
\hline
$\mathrm{QBEC}_{64}$ & 71.29$\pm$0.26 & 64.37$\pm$0.17\\

 \midrule
& \multicolumn{2}{c}{\textbf{Classical Embedding Compression Models}} \\
 \midrule

$\mathrm{BEC}_{384}$ & 72.98$\pm$0.17 & 66.83$\pm$0.12\\
\hline
BEC or $\mathrm{BEC}_{256}$  & 72.83$\pm$0.24 & 66.76$\pm$0.21\\
\hline
$\mathrm{BEC}_{128}$ & 71.93$\pm$0.23 & 66.19$\pm$0.17\\
\hline
$\mathrm{BEC}_{64}$  & 70.84$\pm$0.18 & 64.68$\pm$0.25\\

 \midrule
& \multicolumn{2}{c}{\textbf{Embedding Compression Models with BERT's layers frozen}} \\
 \midrule
 
QBFEC or  $\mathrm{QBFEC}_{256}$ & 72.3$\pm$0.11 & 66.13$\pm$0.14\\
\hline
BFEC or  $\mathrm{BFEC}_{256}$ & 72.66$\pm$0.19 & 66.82$\pm$0.16\\

 \midrule
& \multicolumn{2}{c}{\textbf{Base Models}} \\
 \midrule
 
BU & 73.36 & 67.16\\
\hline
BT & 73.66$\pm$0.15 & 67.22$\pm$0.1\\
\hline
QBT & 73.68$\pm$0.06 & 67.85$\pm$0.03\\
\hline
\end{tabular}
\label{tab:A8.1}
\egroup
\end{center}
\end{tiny}
\vspace{.75em}

\begin{tiny}
\begin{center}
\bgroup
\def\arraystretch{1.5}%
\captionof{table}{Averages and standard errors of performance scores (in NDCG@10) across 20 runs for models using $\mathrm{BERT}_{base}$ as backbone and base models. The scores are scaled by 100.}
\begin{tabular}{c | c c}
 \toprule
 \textbf{Models} & \textbf{TREC19} & \textbf{TREC20} \\

 \midrule
& \multicolumn{2}{c}{\textbf{Quantum-inspired Embedding Compression Models}} \\
 \midrule
QBEC or $\mathrm{QBEC}_{256}$ & 69.81$\pm$0.29 & 66.53$\pm$0.23\\

 \midrule
& \multicolumn{2}{c}{\textbf{Classical Embedding Compression Models}} \\
 \midrule

BEC or $\mathrm{BEC}_{256}$ & 69.17$\pm$0.26 & 65.42$\pm$0.3\\

 \midrule
& \multicolumn{2}{c}{\textbf{Base Models}} \\
 \midrule

BU & 36.57 & 26.13\\
\hline
BT & 69.84$\pm$0.2 & 66.4$\pm$0.2\\
\hline
QBT & 70.73$\pm$0.21 & 67.17$\pm$0.2\\
\hline
\end{tabular}
\label{tab:A8.2}
\egroup
\end{center}
\end{tiny}
\vspace{.75em}

\begin{tiny}
\begin{center}
\bgroup
\def\arraystretch{1.5}%
\captionof{table}{Average performance scores (NDCG@10) with standard errors for classical and quantum-inspired compression models, evaluated across 20 runs using $BERT_{base}$ as backbone. Results are reported for varying training dataset sizes, with scores scaled by 100. 100\% corresponds to the use of 100K training samples.}
\begin{tabular}{c | c | c | c | c | c | c | c | c }
 \toprule
 \multirow{2}{*}{\textbf{Benchmarks}} & \multirow{2}{*}{\textbf{Models}} & \multicolumn{7}{c}{\textbf{\%Training Samples}} \\
 \cmidrule{3-9}
 & & \textbf{1\%} & \textbf{2\%} & \textbf{5\%} & \textbf{10\%} & \textbf{25\%} & \textbf{50\%} & \textbf{100\%} \\
 \midrule

\multirow{2}{*}{\textbf{TREC19}} & BEC & \makecell{45.39\\$\pm$\\0.52} & \makecell{50.14\\$\pm$\\0.4} & \makecell{58.66\\$\pm$\\0.49} & \makecell{64.0\\$\pm$\\0.48} & \makecell{67.8\\$\pm$\\0.2} & \makecell{69.25\\$\pm$\\0.36} & \makecell{69.17\\$\pm$\\0.26}\\
\cmidrule{2-9}
& QBEC & \makecell{49.1\\$\pm$\\0.48} & \makecell{53.08\\$\pm$\\0.48} & \makecell{60.58\\$\pm$\\0.3} & \makecell{65.07\\$\pm$\\0.35} & \makecell{69.15\\$\pm$\\0.26} & \makecell{69.67\\$\pm$\\0.21} & \makecell{69.81\\$\pm$\\0.29}\\
\hline

\multirow{2}{*}{\textbf{TREC20}} & BEC & \makecell{37.91\\$\pm$\\0.39} & \makecell{44.39\\$\pm$\\0.46} & \makecell{51.75\\$\pm$\\0.3} & \makecell{56.39\\$\pm$\\0.45} & \makecell{60.93\\$\pm$\\0.35} & \makecell{63.12\\$\pm$\\0.31} & \makecell{65.42\\$\pm$\\0.3}\\
\cmidrule{2-9}
& QBEC & \makecell{40.51\\$\pm$\\0.42} & \makecell{45.49\\$\pm$\\0.41} & \makecell{53.02\\$\pm$\\0.29} & \makecell{57.91\\$\pm$\\0.34} & \makecell{61.92\\$\pm$\\0.33} & \makecell{65.01\\$\pm$\\0.27} & \makecell{66.53\\$\pm$\\0.23}\\
\hline

\end{tabular}
\label{tab:A8.3}
\egroup
\end{center}
\end{tiny}
\vspace{.75em}

\begin{tiny}
\begin{center}
\bgroup
\def\arraystretch{1.5}%
\captionof{table}{Average performance scores (NDCG@10) with standard errors for classical and quantum-inspired compression models, evaluated across 20 runs by training epoch using $\mathrm{BERT_{base}}$ as backbone. The scores are scaled by 100.}
\begin{tabular}{c | c | c | c | c | c | c | c | c | c | c | c}
 \toprule
 \multirow{2}{*}{\textbf{Benchmarks}} & \multirow{2}{*}{\textbf{Models}} & \multicolumn{10}{c}{\textbf{Training Epochs}} \\
 \cmidrule{3-12}
 & & \textbf{1} & \textbf{2} & \textbf{3} & \textbf{4} & \textbf{5} & \textbf{6} & \textbf{7} & \textbf{8} & \textbf{9} & \textbf{10} \\
 \midrule

\multirow{2}{*}{\textbf{TREC19}} & BEC & \makecell{70.1\\$\pm$\\0.26} & \makecell{70.16\\$\pm$\\0.27} & \makecell{69.13\\$\pm$\\0.24} & \makecell{68.65\\$\pm$\\0.27} & \makecell{68.35\\$\pm$\\0.31} & \makecell{68.08\\$\pm$\\0.32} & \makecell{67.76\\$\pm$\\0.27} & \makecell{67.21\\$\pm$\\0.34} & \makecell{67.14\\$\pm$\\0.33} & \makecell{66.84\\$\pm$\\0.36}\\
\cmidrule{2-12}
& QBEC & \makecell{70.24\\$\pm$\\0.27} & \makecell{70.36\\$\pm$\\0.21} & \makecell{69.97\\$\pm$\\0.27} & \makecell{69.68\\$\pm$\\0.32} & \makecell{69.58\\$\pm$\\0.26} & \makecell{69.41\\$\pm$\\0.26} & \makecell{69.16\\$\pm$\\0.35} & \makecell{68.97\\$\pm$\\0.29} & \makecell{69.09\\$\pm$\\0.29} & \makecell{68.61\\$\pm$\\0.31}\\
\hline

\multirow{2}{*}{\textbf{TREC20}} & BEC &  \makecell{63.74\\$\pm$\\0.31} & \makecell{65.03\\$\pm$\\0.35} & \makecell{65.3\\$\pm$\\0.34} & \makecell{65.07\\$\pm$\\0.24} & \makecell{64.33\\$\pm$\\0.37} & \makecell{64.1\\$\pm$\\0.41} & \makecell{64.13\\$\pm$\\0.34} & \makecell{63.28\\$\pm$\\0.37} & \makecell{63.23\\$\pm$\\0.4} & \makecell{63.07\\$\pm$\\0.34}\\
\cmidrule{2-12}
& QBEC & \makecell{64.26\\$\pm$\\0.23} & \makecell{65.76\\$\pm$\\0.21} & \makecell{66.22\\$\pm$\\0.23} & \makecell{66.47\\$\pm$\\0.19} & \makecell{66.78\\$\pm$\\0.28} & \makecell{66.3\\$\pm$\\0.23} & \makecell{65.86\\$\pm$\\0.23} & \makecell{65.59\\$\pm$\\0.22} & \makecell{65.28\\$\pm$\\0.27} & \makecell{65.27\\$\pm$\\0.31}\\
\hline

\end{tabular}
\label{tab:A8.4}
\egroup
\end{center}
\end{tiny}

\end{appendices}

\end{document}